# Multi-Label Classification with Generative AI Models in Healthcare: A Case Study of Suicidality and Risk Factors

Ming Huang, PhD[1‡*], Zehan Li, MS[1‡], Yan Hu, MS[1], Wanjing Wang, MS[1], Andrew Wen, MS[1], Scott Lane, PhD[2], Salih Selek, MD[2], Lokesh Shahani, MD[2], Rodrigo Machado-Vieira, MD, PhD[2], Jair Soares, MD, PhD[2], Hua Xu, PhD[3], Hongfang Liu PhD[1,2*]

[1]McWilliams School of Biomedical Informatics, The University of Texas Health Science Center at Houston, TX, USA; [2] Faillace Department of Psychiatry & Behavioral Sciences, McGovern Medical School, The University of Texas Health Science at Houston, Houston, TX, USA; [3]Department of Biomedical Informatics and Data Science, School of Medicine, Yale University, New Haven, CT, USA
[‡] Equal contribution
[*] Corresponding authors

**Abstract**

Suicide remains a pressing global health crisis, with over 720,000 deaths annually and millions more affected by suicide ideation (SI) and suicide attempts (SA). Early identification of suicidality-related factors (SrFs), including SI, SA, exposure to suicide (ES), and non-suicidal self-injury (NSSI), is critical for timely intervention. While prior studies have applied AI to detect SrFs in clinical notes, most treat suicidality as a binary classification task, overlooking the complexity of co-occurring risk factors. This study explores the use of generative large language models (LLMs), specifically GPT-3.5 and GPT-4.5, for multi-label classification (MLC) of SrFs from psychiatric electronic health records (EHRs). We present a novel end-to-end generative MLC pipeline and introduce advanced evaluation methods, including label-set–level metrics and a multi-label confusion matrix for error analysis. Fine-tuned GPT-3.5 achieved top performance with 0.94 partial-match accuracy and 0.91 F1 score, while GPT-4.5 with guided prompting showed superior performance across label sets, including rare or minority label sets, indicating a more balanced and robust performance. Our findings reveal systematic error patterns, such as the conflation of SI and SA, and highlight the models' tendency toward cautious over-labeling. This work not only demonstrates the feasibility of using generative AI for complex clinical classification tasks but also provides a blueprint for structuring unstructured EHR data to support large-scale clinical research and evidence-based medicine.

**Significance Statement**

Suicide prevention remains a global health priority. This study pioneers the use of generative large language models (LLMs), such as GPT-3.5 and GPT-4.5, for the identification of SrFs from unstructured clinical text via multi-class classification. By introducing novel evaluation and error analysis methods, we reveal how these models can accurately and robustly identify nuanced risk profiles, even in imbalanced datasets. Our approach transforms narrative electronic health records into structured data, enabling large-scale medical research and advancing evidence-based medicine.

## Introduction

Suicide is one of the leading causes of death worldwide, presenting a critical public health concern [1]. World Health Organization (WHO) estimated that more than 720,000 people died by suicide globally each year [2,3]. In the United States, Centers for Disease Control and Prevention (CDC) reported over 49,000 lives lost to suicide in 2022 [4]. The data from CDC also indicates around 1.6 million suicide attempts and 13.2 million suicide thoughts, highlighting a prevalent issue of suicidality [4].

Mitigating suicide risk are imperative to address the placed emotional, physical, economic, and societal challenges [3,5]. Extensive research has demonstrated that the risk of suicide is significantly elevated among individuals with a history of suicidality and related factors (SrFs) such as suicidal ideation (SI), and suicide attempts (SA), exposure to suicide (ES), and non-suicidal self-injury (NSSI) [6-8]. These findings underline the critical need for early identification of these factors to guide timely and lifesaving interventions. Trained mental health providers identify and interpret signs of suicidality, even when individuals may deny, minimize, or be evasive about their conditions. These observations are documented in electronic health records (EHRs), particularly within clinical notes, providing a valuable resource for studying suicidality.

Multiple studies have demonstrated the effectiveness of artificial intelligence (AI) techniques such as natural language processing (NLP) or machine learning in detecting SrFs from clinical notes in recent years [9-13]. A common limitation of these studies, however, is their focus on a single aspect of suicidality – identifying either general suicidality or a specific type (SI or SA) as highlighted in a recent systematic review where all 35 included studies on suicidal behavior predictions during 2016-2021 were implemented as binary classification tasks [14]. In real-world psychiatric services, patients often present with multiple stages of suicidality in their lives and the corresponding clinical notes document their various SrFs.

Recently, large language models (LLMs) have become transformative tools in AI, leveraging massive datasets and advanced architectures (e.g., Transformers) to obtain pre-trained models through either encoding or decoding techniques that can be fine-tuned for domain-specific applications [15,16]. A study previously investigated the use of bidirectional encoder representations from transformer (BERT) models for identifying SrFs in clinical notes through multi-label classification (MLC) with RoBERTa superior than other BERT models [17]. Given the dominance of generative AI models, in particular, generative pre-trained transformer (GPT) models [16,18], here we explore the use of GPT models for identifying SrFs through multi-label classification (MLC) not only because the use of GPT for detecting SrFs is underexplored [17,19] but also because the application and evaluation of MLC for clinical information extraction is limited studied.[20] Thus, we propose a novel performance and error analysis method, aiming to provide insights about the model's strengths and weaknesses, which can be readily adapted to other MLC tasks and health information detection problems. This MLC approach also pave the way for converting unstructured information to structured formats, enabling observational studies for mental health and other healthcare domains that advance evidence-based medicine.

This study leverages generative GPT models to perform MLC tasks via a case study by classifying SrFs from real-world psychiatric notes, and it makes four contributions:
1. End-to-end generative MLC pipeline with LLMs. We demonstrate that fine-tuned GPT-3.5 and guideline prompted GPT-4.5 can achieve top accuracy of 0.93-0.94 and F1 scores of 0.88–0.91 on a MLC task with four labels despite severe class imbalance.
2. Label-set–level evaluation. Beyond conventional micro/macro metrics, we introduce exact- and partial-match label-set metrics that quantify performance on clinically meaningful label combinations, capturing errors that standard label-level statistics obscure.
3. Multi-label confusion-matrix error analysis. We extend the classical confusion matrix to the power-set setting, enabling granular inspection of hallucination (false-positive) versus omission (false-negative) patterns. For instance, we reveal that nearly half of SA-only notes are mislabelled as SI & SA, spotlighting a systematic tendency to conflate ideation with attempts.

4. Blueprint for structuring narrative EHR data. Our workflow enables the conversion of unstructured clinical text into structured tables, facilitating large-scale observational studies and downstream causal analyses in healthcare research.

**Results**

In the following, we present our results on the comparison of GPT models with relevant training strategies for identifying multiple SrFs and the adoption of a novel multi-facet performance and error analysis method for MLC task, aiming to provide insights of the model behavior for identifying SrFs from clinical notes.

**Model Performance Analysis**

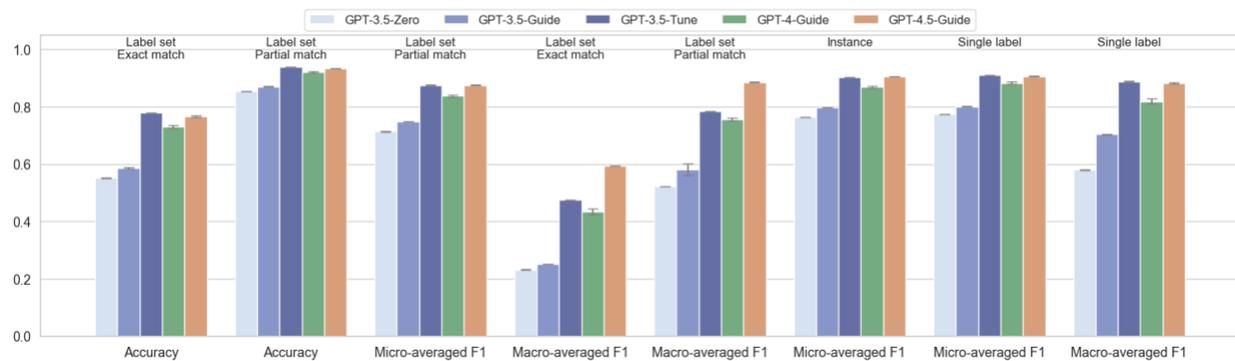

**Figure 1**. Overall model performance in terms of accuracy and micro- and macro-averaged F1 scores. GPT-3.5-Zero/-Guide/-Tune: GPT-3.5 with zero-shot learning, zero-shot learning and a guideline, or fine-tuning; GPT-4/4.5-Guide: GPT-4 or 4.5 with zero-shot learning and a guideline

As illustrated in **Figure 1**, GPT-3.5-Tune achieved the highest accuracy, with 0.78±0.00 in exact match and 0.94±0.00 in partial match. GPT-3.5-Guide (0.59±0.00 in exact match and 0.87±0.00 in partial match) outperformed GPT-3.5-Zero (0.55±0.00 in exact match and 0.85±0.00 in partial match), reflecting the performance improvement via a guideline. For zero-shot learning with a guideline, the performance was continuously improved with the advancement of models as demonstrated by GPT-4-Guide (0.73±0.01 in exact match and 0.92±0.00 in partial match) and GPT-4.5 (0.77±0.00 in exact match and 0.93±0.00 in partial match). The performance of GPT-4.5-Guide was slightly lower than GPT-3.5-Tune by 0.01. The accuracy difference between the best and worst models decreased from 0.23 in exact match to 0.09 in partial match.

For micro-averaged F1 scores, similar performance patterns were observed for these five models across multiple metrics including label sets, instances, and single labels. GPT-3.5-Tune demonstrated the highest F1 scores (0.88-0.91) for most metrics. However, for the instance-based metric, the micro-averaged F1 score of GPT-4.5-Guide (0.91±0.00) was slightly higher than GPT-3.5-Tune by 0.01. For macro-averaged F1 scores, the performance patterns were also similar to accuracies and micro-averaged F1 scores for GPT-3.5 and GPT-4. However, GPT-4.5-Guide showed the highest macro-averaged F1 score of 0.59±0.00 (exact match) and 0.89±0.00 (partial match), which was significantly better than those of GPT-3.5-Tune by 0.10-0.11.

**Single Label Performance Analysis**

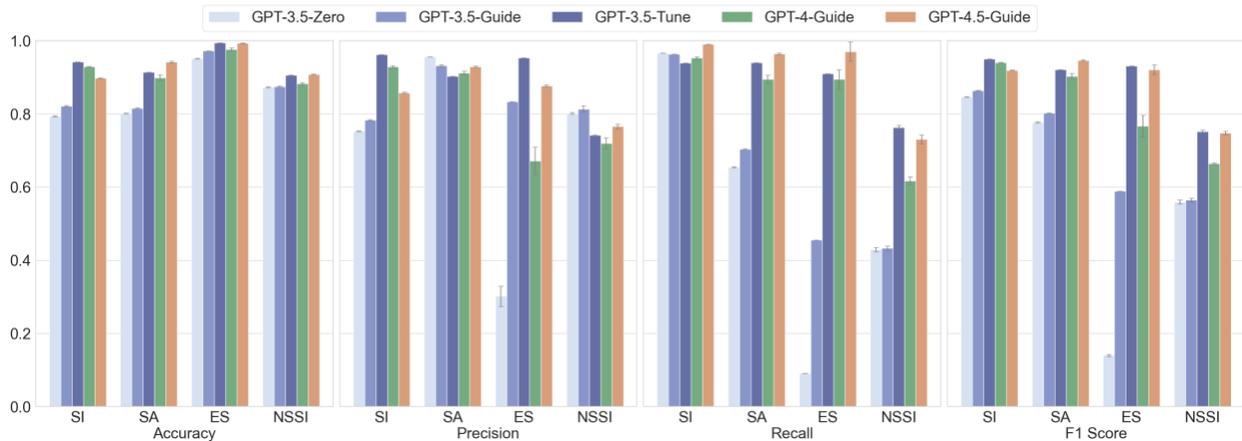

**Figure 2**. Model performance for the four single labels - SI, SA, ES, and NSSI. SI: suicidal ideation; SA: suicide attempts; ES: exposure to suicide; NSSI: non-suicidal self-injury.

In terms of accuracies and F1 scores, either GPT-3.5-Tune or GPT-4.5-Guide demonstrated top performance for different labels as shown in **Figure 2**. For example, for SI (or SA), GPT-3.5-Tune (or GPT-4.5-Guide) outperformed the others, achieving an accuracy and F1 score of 0.94-0.95. GPT-3.5-Tune and GPT-4.5-Guide had comparable accuracy of 0.99 for ES (or 0.91 for NSSI) and F1 score of 0.92-0.93 for ES (or 0.75 for NSSI). Across all the labels, GPT-3.5-Guide consistently outperformed GPT-3.5-Zero and GPT-4-Guide consistently outperformed GPT-3.5-Guide. GPT-4.5-Guide performed better than GPT-4-Guide for most labels except SI.

Among the four labels, SI and SA consistently showed the best performance across all models. For ES, we observed consistently highest accuracies of 0.96-0.99 across all models, but a large variation in F1 scores of 0.14-0.93 for these models. Notably, GPT-3.5-Zero had the lowest F1 score of 0.14±0.00 and the F1 scores of GPT-3.5 was improved significantly via a guideline (0.59±0.00) and fine-tuning (0.93±0.00). Compared with GPT-3.5-Guide, the model performance for ES was further improved with GPT-4-Guide (0.77±0.03) and GPT-4.5-Guide (0.93±0.02). For the best models (GPT-3.5-Tune or GPT-4.5-Guide), NSSI is the most challenging label with the lowest accuracy of 0.91 and F1 score of 0.75.

**Label Set Performance Analysis**

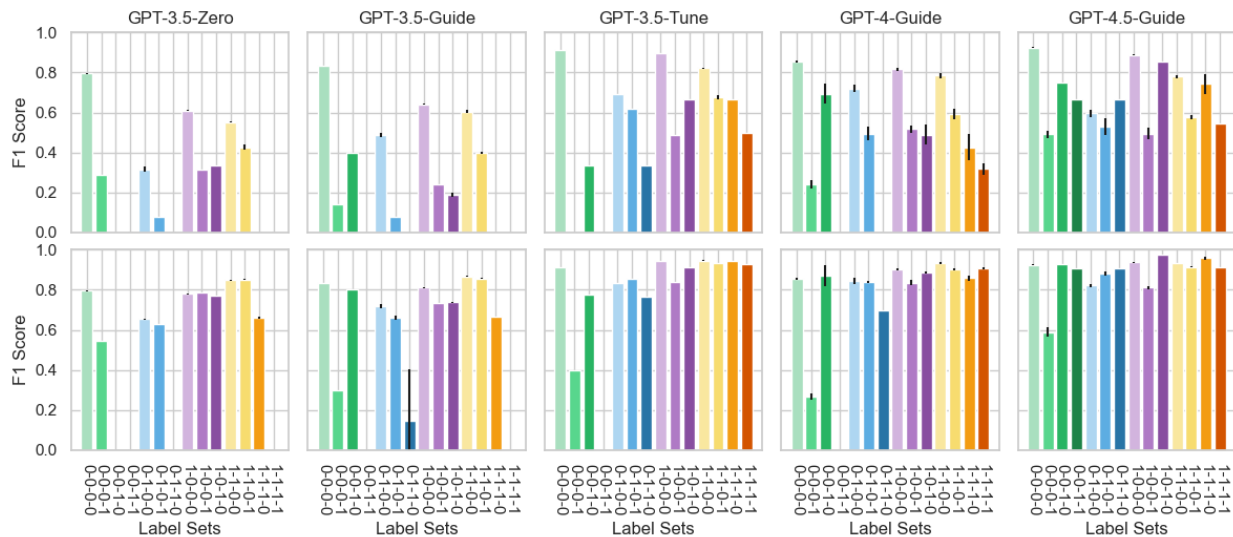

**Figure 3**. Model performance for the 14 label sets with exact match (top) and partial match (bottom) in terms of F1 scores. See additional metrics in the **supplementary document**.

We illustrate the model performance for each label set in exact match and partial match in **Figure 3**. For the exact match performance, the models generally performed well on the four high-frequency label sets (*-*-0-0), while their performance decreased for most medium- and low-frequency label sets involving ES and NSSI. The best-performing models varied across the label sets, with F1 scores ranging from 0.49 to 0.92. GPT-4.5-Tune achieved the best performance in 8 label sets, including 1-0-1-0 and 0-0-0-0 (F1=0.86-0.92), 1-1-1-0 and 0-0-1-0 (F1=0.74-0.75), 0-0-1-1 and 0-1-1-0 (F1=0.67), and 0-0-0-1 and 1-1-1-1 (F1=0.49-0.55). GPT-3.5-Tune led in 4 label sets: 1-1-0-0 and 1-0-0-0 (F1=0.82-0.90) and 0-1-0-1 and 1-1-0-1 (F1=0.62-0.68). For the remaining two label sets, GPT-4-Guide outperformed others for 0-1-0-0 (F1=0.72±0.02) and 1-0-0-1 (F1=0.52±0.02).

Compared with exact match, all models showed considerable improvement in partial match across nearly all label sets except 0-0-0-0. The best-performing models for 11 of 14 unique label sets in exact match are remains the same. The best F1 scores of the 13 unique label sets are larger than 0.84, 8 of which are more than 0.90. Among the 70 F1 scores of 14 label sets predicted by the 5 models, 3, 18, and 40 F1 scores increased by over 60%, 40%, and 20%, respectively, compared with their exact match scores. Three F1 scores of 0 in exact match increased their scores to over 0.66 in partial match, including 1-1-1-0 by GPT-3.5-Zero and GPT-3.5-Guide and 0-1-1-0 by GPT-4-Guide. The findings suggests that even when the models failed to perform well in exact matches, they were still capable of capturing parts of the label sets correctly, which is critical in clinical decision-making contexts where some predictive accuracy is better than none.

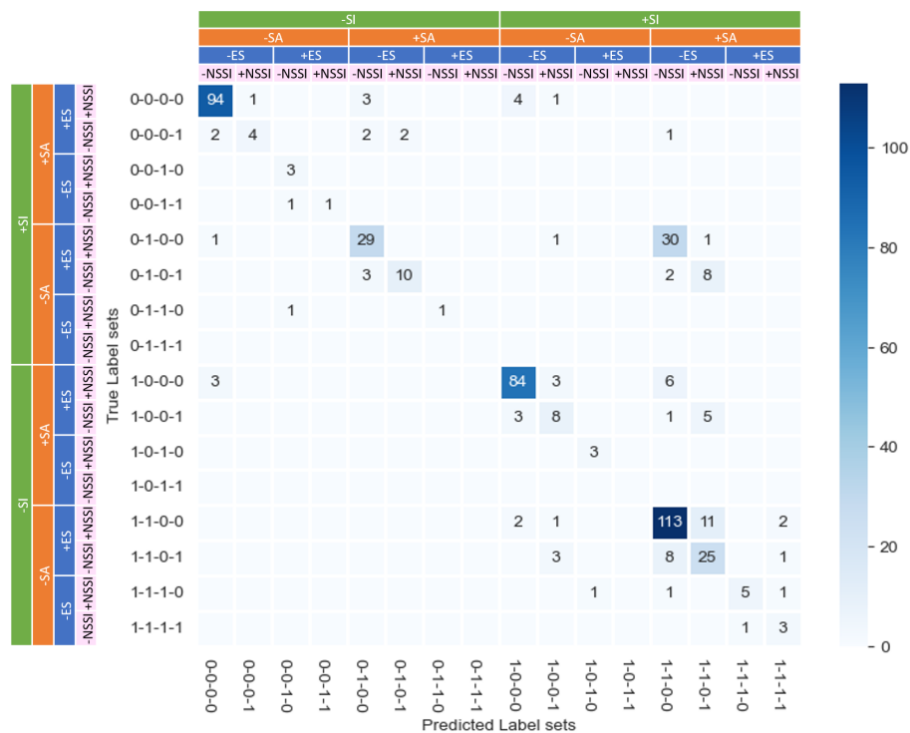

**Figure 4**. Confusion matrix of multi-label classification (MLC) with GPT-4.5-Guide along with both binary and semantic codes

**Error Analysis and Visualization**

**Figure 4** presents a 16x16 confusion matrix of MLC with GPT-4.5-Guide along with both binary and semantic codes for error analysis. The confusion matrix can provide error analysis in three levels:

(i) Overall model. Out of the 500 cases, we observed 86 errors (80 hallucinations) in the top-right corner of the confusion matrix, compared to 31 errors (28 omissions) in the bottom-left corner. More hallucinations in the top-right corner indicate that the model tends to over-predict labels (false positives). Fewer errors in the bottom-left corner suggest the model seldom under-predicts complexity.

(ii) Label set. Among the four high-frequency label sets, 0-1-0-0 (SA-only) had the lowest F1 scores of 0.60±0.02 in exact match and 0.82±0.01 in partial match (See **Figure 3**). 30 out of 62 true 0-1-0-0 cases were misclassified as 1-1-0-0 (SI&SA). The errors show that the model is challenging to separate SA-only from SI&SA cases. Additionally, GPT-4.5-Guide had F1 scores of 0.49±0.02 in exact match and 0.59±0.02 in partial match, respectively, for 0-0-0-1 (NSSI-only). 7 out of 11 true 0-0-0-1 cases were misclassified and 5 of the 7 errors contain a hallucinatory SA label (2 SA-only, 2 SA&NSSI, and 1 SI&SA predictions), suggesting the model is prone to falsely associate NSSI with SA for 0-0-0-1.

(iii) Single label. The confusion matrix also enables the error analysis of each single label or a group of label sets containing the single label. For example, the model made far more hallucination errors (48) than omission errors (3) for SI among the 294 cases. Of the hallucination errors, 79% (38/48) arose when the model misclassified 0-1-0-* as 1-1-0-*. This shows the model has difficulty separating SA-only cases from SI&SA cases, often adding an unwarranted SI label.

**Discussion**

According to the interpersonal theory of suicide, suicidality exists on a continuum from suicidal ideation to completed suicide [21,22]. Recognizing this continuum is crucial for accurately predicting suicide risks and informing treatment decisions. Patients admitted to psychiatric services often present with varying historical and present stages of suicidality, all documented in clinical notes. The complex nature of suicidal behavior highlights the importance of developing AI solutions capable of identifying multiple SrFs within unstructured clinical narratives to enhance intervention strategies. In this study, we investigated MLC of SrFs in IPE notes using LLMs including GPT-3.5, GPT-4, and GPT-4.5 via multiple learning strategies such as zero-shot learning with and without a guideline and fine-tuning. The model performance was evaluated at multiple levels including model level, single label, and label-set level.

The analysis of overall model performance indicates that the fine-tuned GPT-3.5 achieved the best overall performance, with accuracies of 0.78 (exact match) and 0.94 (partial match) and micro-averaged F1 scores of 0.88-0.91 across instances, single labels, and label sets. Zero-shot GPT-4.5 with guided prompting demonstrated comparable effectiveness, with accuracies of 0.77 (exact match) and 0.93 (partial match). These outcomes suggest the promise of generative AI models to accurately identify SrFs from IPE notes. Comparisons across learning strategies (GPT-3.5-Zero, GPT-3.5-Guide, GPT-3.5-Tune) underscored the advantages of fine-tuning over guided prompting and guided prompting over pure zero-shot learning. Comparative analyses among different GPT models with the same guided prompting (GPT-3.5-Guide, GPT-4-Guide, GPT-4.5-Guide) revealed incremental improvements associated with model advancement.

Inspired by instance-based metrics which can identify the challenging instances, we proposed the novel label-set level metrics including both exact match and partial match to examine how the model performs across label sets. These metrics can measure overall model performance on label sets and provide insights into particularly challenging label sets. For example, compared with GPT-3.5-Tune, GPT-4.5-Guide has significantly superior performance in macro-averaged F1 score (by 10-11%). It suggests an important difference: while GPT-3.5-Tune was marginally more accurate on the most frequent or dominant label sets, GPT-4.5-Guide demonstrates a substantially better capability to correctly classify instances across all label sets, including rare or minority label sets, indicating a more balanced and robust performance. We also observed that the best-performing models varied across the label sets. GPT-4.5-Tune achieved the best performance in 8 label sets, GPT-3.5-Tune in 4-5 label sets and GPT-4-Guide in 1-2 label sets. All the models failed to perform well for 0-0-0-1 (NSSI-only). The highest F1 scores of 0-0-0-1 were 0.49-0.59 achieved by GPT-4.5-Guide. A novel error analysis of MLC was developed to provide insights into the specific error patterns about models, label sets, and single labels. For instance, more hallucinations in the top-right corner of confusion matrix of GPT-4.5-Guide suggests a tendency toward sensitivity or caution, labeling cases as more severe or complex than they actually are. Conversely, fewer omissions in the bottom-left corner suggest the model rarely misses labels completely. We also found that GPT-4.5-Guide was prone to falsely associate NSSI with SA for processing 0-0-0-1. Understanding these error patterns is crucial for improving model robustness and refining its predictions for complex MLC tasks.

MLC is a challenging task for traditional machine learning and deep learning models [23], due to the high dimensionality of label sets and the issue of data imbalance [24-26]. Various sophisticated multi-label learning methods have been developed to address these challenges [25,26]. Among these methods, binary relevance method is often deployed to transform MLC into multiple binary classification, particularly for healthcare applications by considering its simplicity and ease of implementation [27]. However, multiple binary classification treats each single label individually and ignores the relevance among these labels, achieving suboptimal performance. As shown in a previous work with RoBERTa, the single MLC outperformed multiple binary classification by 3% in identifying SrFs [17]. This work demonstrated the effectiveness of generative AI models in identifying multiple SrFs from IPE notes. Although we experimented MLC for identifying four SrFs in IPE notes, the work could be easily extended to detect more labels and identify other health information from various texts, even beyond the healthcare domain. Additionally, MLC can be naturally adapted for text-to-table tasks which convert unstructured clinical notes into structured formats [28], ready for data integration, statistical analysis, and observational studies. Traditionally, the structured health information was often extracted from unstructured health text by using information extraction methods (e.g., named entity recognition and relation extraction) and then was normalized for integration and analysis [29-33]. Compared with information extraction technique, MLC reduces the task complexity by ignoring the specific phrase to be extracted and its location in the given text. MLC has the potential to open a new avenue for text-to-table tasks, enabling researchers and clinicians to perform observational studies that advance evidence-based medicine.

**Materials and Methods**

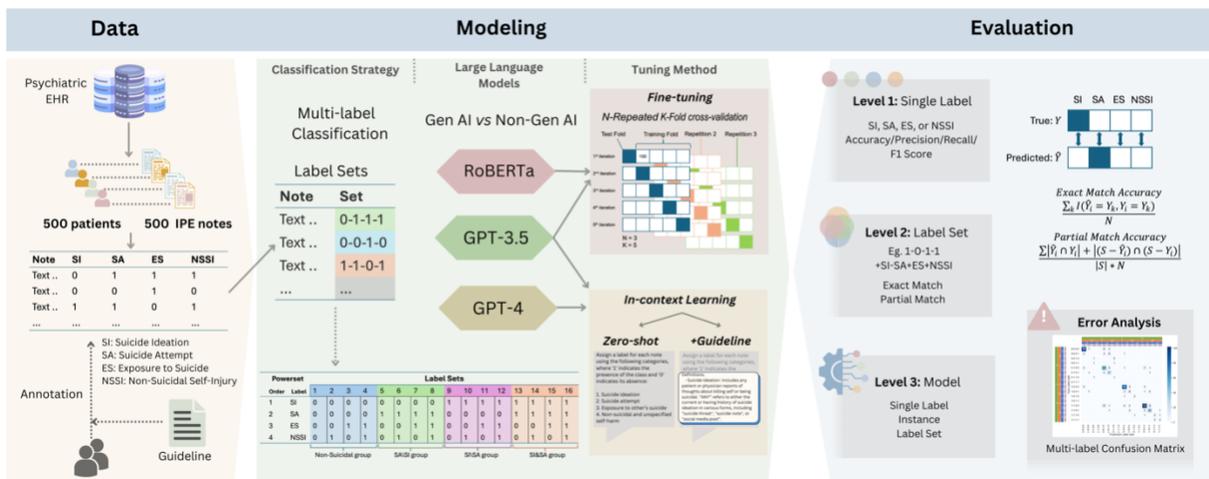

**Figure 5**. Workflow of multi-label classification (MLC) of suicidality and related factors (SrFs) from initial psychiatric evaluation (IPE) notes using large language models (LLMs)

**Figure 5** illustrates the workflow to identify multiple SrFs from initial psychiatric evaluation (IPE) notes using MLC with large language models (LLMs). Specifically, an annotated corpus of 500 IPE notes was used to develop and validate MLC algorithms for SrF detection. Each note was labeled with a set of four binary codes (0 or 1), indicating the presence (1) or absence (0) of mentions for each category of SrFs including SI, SA, ES, and NSSI. For generative AI models such as GPT-3.5, GPT-4, and GPT-4.5, we designed prompts with and without a guideline for zero-shot learning to instruct them to generate a set of four binary codes (0 or 1) for MLC of the four SrFs. GPT-3.5 was further fine-tuned by feeding pairs of IPE notes as input and sets of four labels as output through 5-fold cross-validation. Each zero-shot learning and fine-tuning experiment was repeated for 3 times to ensure the robustness and generalizability of each model across experimental setups and/or data splits. We evaluate the model performance using multiple metrics across different dimensions: label level, label-set level, and model level. Notably, we created novel label-set-level metrics based on exact match and partial match, which enable the evaluation of the model performance for various combination of labels. We also propose a new error analysis approach to examine the model error for MLC. In the following sections, we will discuss each of these components in **Figure 5**, including study dataset, MLC methods, experiments and settings, performance evaluation, and error analysis in more details.

**Study Dataset**

The study dataset includes 500 IPE notes from the Electronic Health Record (EHR) database of a safety-net hospital for inpatient psychiatric care. This hospital offers world-class expertise and professional care in psychiatry and behavioral health services. The IPE notes contain a detailed assessment of each patient's mental health history, current symptoms, and overall functioning during their initial admission. The evaluation is critical for diagnosing psychiatric disorders, including assessing suicide tendencies, to ensure timely intervention.

The 500 IPE notes were annotated by two annotators in our research team by following the TRUST process, a standard data abstraction and annotation process [34]. Each note was labeled with a set of four binary codes (0 or 1), indicating the presence (1) or absence (0) of mentions for each category of SrFs in the following order: "Suicidal Ideation" (SI), "Suicide Attempts" (SA), "Exposure to Suicide" (ES), and "Non-Suicidal Self-Injury" (NSSI). "Suicidal Ideation" (SI) refers to thoughts about killing oneself, as mentioned by patients or physicians in the notes. "Suicide Attempts" (SA) denotes self-injurious behavior with the intent to die, which did not result in fatality. "Exposure to Suicide" (ES) involves suicidality (SI, SA, or suicide death) experienced by individuals other than the patient, such as family members or friends. "Non-Suicidal Self-Injury" (NSSI) includes deliberate self-inflicted harm to body tissue without the intent to die and not socially sanctioned, as well as non-suicidal self-injurious thoughts. The Inter-Annotator Agreement (IAA) before adjudication was assessed with Cohen's Kappa (κ) score, resulting values of 0.98 for SI, 1.00 for SA, 0.98 for ES, and 0.90 for NSSI, averaging κ = 0.95, indicating near-perfect agreement (0.81–1.00) [35]. Disagreements were adjudicated and resolved through discussion with a panel of 4 board-certified psychiatrists (RV, SS, LS, and JS).

The four labels (SI, SA, ES, and NSSI) generate 16 possible combinations or label sets, which can be represented as a binary code (e.g., 1-0-1-0) in the order of SI, SA, ES and NSSI, a semantic code (e.g., +SI-SA+ES-NSSI), a math set (e.g., {SI, ES}), or a textual code (e.g., SI&ES). We use these notations together with wild cards to denote a group of label sets. For example, the 16 label sets could be categorized into four groups based on SI and SA: non-suicidal group (0-0-*-* or -SI-SA*), SA\SI group (0-1-*-* or -SI+SA*), SI\SA group (1-0-*-* or +SI-SA*), and SI&SA group (1-1-*-* or +SI+SA*). See **Table S1** in the **supplementary document** for more details. The binary code representation will be primarily used for the sequential analysis and discussion, complemented with other notations for better explanation and understanding.

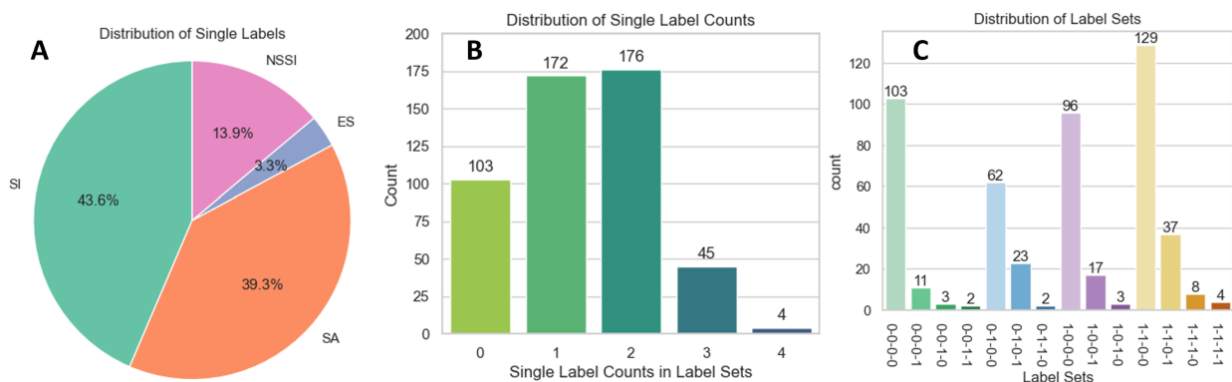

**Figure 6**. Distributions of single labels and label sets in the annotated corpus

The annotated corpus of 500 IPE notes includes 675 single labels of SrFs with a distribution of 294 (43.6%) SI, 265 (39.3%) SA, 22 (3.3%) ES, and 94 (13.9%) NSSI (See **Figure 6A**). Out of the 16 possible label sets, we identified 14 unique label sets in the sampled dataset, missing two label sets of 0-1-1-1 and 1-0-1-1. As shown in **Figure 6B**, 90.2% of the 500 label sets have 0-2 labels, and only 9.0% and 0.8% label sets contain 3 and 4 labels, respectively. **Figure 6C** illustrates the frequencies of the 14 identified label sets within four groups based on SI and SA: non-suicidal group, SA\SI group, SI\SA group, and SI&SA group. Each label-set group has a high-frequent label set ending with 0-0 (*-ES-NSSI) such as 0-0-0-0 (None) in the non-suicidal group, 0-1-0-0 in the SA group, 1-0-0-0 in the SI group, and 1-1-0-0 in the SI&SA group. Within each label-set group, we also observe a moderate-frequent label set ending 0-1 (*-ES+NSSI), and 1-2 low-frequent label sets ending with 1-0 (*+ES-NSSI) and/or 1-1 (*+ES+NSSI).

**Multi-label Classification (MLC)**

To identify the multiple SrFs at the document level, we employed a MLC approach with LLMs. MLC enables the simultaneous identification of multiple labels that are not mutually exclusive [24]. We leveraged three Generative Pre-trained Transformer (GPT) models for MLC tasks, including GPT-3.5, GPT-4, and GPT-4.5. [36,37] We designed a prompt with or without a guideline (See **supplementary document** for the prompts) to instruct these GPT models to generate a set of four binary codes (0 or 1), indicating the presence (1) or absence (0) of the four SrFs in the following order: SI, SA, ES, and NSSI. In addition, we further fine-tuned the GPT-3.5 model by feeding pairs of IPE notes as input and sets of four labels as output. We performed two experiments for MLC tasks with LLMs: (1) three different learning strategies for GPT-3.5 – zero-shot learning (GPT-3.5-Zero), zero-shot learning with a guideline (GPT-3.5-Guide), and fine-tuning (GPT-3.5-Tune); (2) three different GPT models with the guided prompting – GPT3.5-Guide, GPT-4-Guide, and GPT-4.5-Guide (See **supplementary document** for more details about experiment settings).

**Performance Evaluation**

To assess the model performance of MLC, we proposed and employed a series of metrics that capture model behaviors across different dimensions: (1) label-level metrics, (2) label-set-level metrics, and (3) model-level metrics. The three types of metrics complement each other to provide a multifaced assessment of model performance. Label-level metrics quantify the performance of the model for each individual label. Label-set-level metrics proposed here evaluate the performance of the model for each label set. Model-level metrics provide an overall evaluation of the model performance across all instances, labels or label sets. We will describe the label-set-level metrics and relevant model-level metrics in the following sections and provides label-level and instance-level metrics and relevant model-level metrics in the **supplementary document**.

**Label-set-level metrics**

We propose novel label-set-level metrics, especially, partial match metrics, to quantify MLC performance for each combination of labels, or label set. Unlike label-level metrics, which assess performance for each label independently, the label-set-level metric evaluates the performance of the entire label set as a single unit. This approach captures the performance of predicting the correct combination of labels. The label-set-level metrics are divided into two categories: exact match and partial match. Exact match is the most stringent metric, requiring all labels in a set to be predicted correctly for a given instance to be considered fully accurate. In contrast, partial match is a more lenient metric where some, but not all, labels in a set are correctly predicted, providing a more nuanced evaluation of model performance.

For a set of $L$ unique single labels, $S = \{y_1, y_2, ..., y_L\}$ where $L = |S|$, the label powerset $P(S) = \{\emptyset, \{y_1\}, \{y_2\}, ..., \{y_1, y_2, ..., y_L\}\}$ and the number of elements in the powerset $P(S)$ is $M = |P(S)| = 2^L$. For a dataset with $N$ instances, each instance $i$ is associated with a true label set $Y_i$ and a predicted label set $\hat{Y}_i$, where $i = \{1, 2, ..., N\}$, $Y_i \in P(S)$, and $\hat{Y}_i \in P(S)$. These $N$ true label sets $\{Y_i\}$ and predictions $\{\hat{Y}_i\}$ are grouped into $M$ unique label sets $\{Y_k\}$ or $\{\hat{Y}_k\}$, where $k = \{1, ..., M\}$, $Y_k \in P(S)$, and $\hat{Y}_k \in P(S)$.

***Label-set-level metrics based on exact match***. To calculate the exact match metrics at the label-set level, we treat each unique combination of labels, or label set, as a distinct superclass, transforming the evaluation of a MLC problem at the label-set level as that of a multi-class classification problem in the superclass level. For each unique label set $Y_k$, we calculated precision, recall, and F1-score as these of a multi-class classification problem by computing true positive (TP), false positive (FP), and false negative (TN) for the superclass $Y_k$.

$$TP_k = \sum I(\hat{Y}_i = Y_k, Y_i = Y_k)$$

$$FP_k = \sum I(\hat{Y}_i = Y_k, Y_i \neq Y_k)$$

$$FN_k = \sum I(\hat{Y}_i \neq Y_k, Y_i = Y_k)$$

$$Precision_k = \frac{TP_k}{TP_k + FP_k}$$

$$Recall_k = \frac{TP_k}{TP_k + FN_k}$$

$$F1_k = 2\frac{Precision_k * Recall_k}{Precision_k + Recall_k}$$

***Label-set-level metrics based on partial match.*** For the partial match at the label-set level, we measure the match degree between the predicted label set and the true label set, rather than the strict match (1) or mismatch (0) in the exact match metrics. We introduced partial true positive (PTP) from false positives (FP) in exact match for calculating precision in partial match and PTP from false negatives (FN) in exact match for calculating recall in partial match.

$$TP_k = \sum I(\hat{Y}_i = Y_k, Y_i = Y_k)$$

$$FP_k = \sum I(\hat{Y}_i = Y_k, Y_i \neq Y_k)$$

$$FN_k = \sum I(\hat{Y}_i \neq Y_k, Y_i = Y_k)$$

$$PTP_k(FP) = \sum_{\hat{Y}_i = Y_k, Y_i \neq Y_k} \frac{|\hat{Y}_i \cap Y_i|}{|\hat{Y}_i \cap Y_i| + (|\hat{Y}_i \setminus Y_i| + |Y_i \setminus \hat{Y}_i|)/2}$$

$$PTP_k(FN) = \sum_{\hat{Y}_i \neq Y_k, Y_i = Y_k} \frac{|\hat{Y}_i \cap Y_i|}{|\hat{Y}_i \cap Y_i| + (|\hat{Y}_i \setminus Y_i| + |Y_i \setminus \hat{Y}_i|)/2}$$

$$Precision_k = \frac{TP_k + PTP_k(FP)}{TP_k + FP_k}$$

$$Recall_k = \frac{TP_k + PTP_k(FN)}{TP_k + FN_k}$$

$$F1_k = 2\frac{Precision_k * Recall_k}{Precision_k + Recall_k}$$

**Model-level metrics**

The overall model performance is evaluated by aggregating these instance-level, label-level, and label-set-level metrics (including both exact match and partial match) through the micro and macro averaging strategies. We describe the label-set-level aggregation in the following sections and provided the label-level and instance-level aggregation in the **supplementary document**.

***Micro averaging of label-set-level metrics based on exact match***. The micro averaging based on exact match in MLC aggregates the contributions of all label sets across all instances in terms of exact match for computing overall performance scores. It treats all instances equally and calculates micro-averaged precision, recall, and F1-score by summing the true positive (TP), false positive (FP), and false negative (FN) of each label set. The micro averaged precision, recall, and F1 score are identical, corresponding to exact match accuracy or exact match ratio [24].

$$TP_{micro} = \sum_{k} TP_k$$

$$FP_{micro} = \sum_{k} FP_k$$

$$FN_{micro} = \sum_{k} FN_k$$

$$Precision_{micro} = \frac{TP_{micro}}{TP_{micro} + FP_{micro}}$$

$$Recall_{micro} = \frac{TP_{micro}}{TP_{micro} + FN_{micro}}$$

$$F1_{micro} = 2 \frac{Precision_{micro} * Recall_{micro}}{Precision_{micro} + Recall_{micro}}$$

$$Accuracy = \frac{\sum_{k} I(\hat{Y}_i = Y_k, Y_i = Y_k)}{N} = Precision_{micro} = Recall_{micro} = F1_{micro}$$

***Micro averaging of label-set-level metrics based on partial match***. Similar to the micro averaging based on exact match in MLC, the micro averaging based on partial match treats all instances equally and calculates micro-averaged precision, recall, and F1-score by summing the true positive (TP), partial true positive (PTP), false positive (FP), and false negative (FN) of each label set. The micro averaged precision, recall, and F1 score are identical. Model accuracy is calculated by summing the accuracy of each instance and dividing by N, the total number of instances. The accuracy calculated from the label-set-level metrics in partial match is identical to theses calculated at the instance level and single-label level.

$$TP_{micro} = \sum_{k} TP_k$$

$$FP_{micro} = \sum_{k} FP_k$$

$$FN_{micro} = \sum_{k} FN_k$$

$$PTP_{micro}(FP) = \sum_{k} PTP_k(FP)$$

$$PTP_{micro}(FN) = \sum_k PTP_k(FN)$$

$$Precision_{micro} = \frac{TP_{micro} + PTP_{micro}(FP)}{TP_{micro} + FP_{micro}}$$

$$Recall_{micro} = \frac{TP_{micro} + PTP_{micro}(FN)}{TP_{micro} + FN_{micro}}$$

$$F1_{micro} = 2\frac{Precision_{micro} * Recall_{micro}}{Precision_{micro} + Recall_{micro}} = Precision_{micro} = Recall_{micro}$$

$$Accuracy_i = \frac{|\hat{Y}_i \cap Y_i| + |(S - \hat{Y}_i) \cap (S - Y_i)|}{|S|}$$

$$Accuracy = \frac{\sum_i^N Accuracy_i}{N}$$

***Macro averaging of label-set-level metrics***. The macro averaging based on exact match or partial match in MLC evaluates the model performance by calculating the performance metrics (precision, recall, F1-score) for each label set and computing the average of these metrics across all unique label sets. Unlike micro averaging, which weighs all instances equally, macro averaging treats each unique label set equally, regardless of its frequency in the dataset.

$$Metric_{macro} = \frac{1}{M}\sum_{k=1}^{M} Metric_k$$

where $Metric_k$ denotes exact or partial match precision, recall, or F1-score for the k-th unique label set $Y_k$.

**Error Analysis**
We introduced an innovative analytic method based on the label powerset P(S) rather than single labels S for the error analysis of MLC. Specifically, this approach transformed the MLC problem of the single labels, S, into a multi-class classification task involving all label sets, P(S), treating each label set as a unique class. The error analysis of MLC could be performed by generating a MxM confusion matrix of the label sets together with both binary and semantic code.

In this study, we demonstrated the error analysis of MLC with the best-performing model (GPT-4.5-Guide) by generating a 16x16 confusion matrix of the label sets together with both binary and semantic code (See **Figure 4**). The confusion matrix reveals two primary types of predictive errors produced by generative AI models: hallucination and omission error. Hallucination errors occur when the model incorrectly predicts labels in the predicted label set (e.g., SI in +SI+SA-ES-NSSI or 1-1-0-0) that are not present in the true label set (e.g., -SI+SA-ES-NSSI or 0-1-0-0), resulting false positives. Conversely, omission errors occur when the model fails to predict single labels in the predicted label set (e.g., SA in +SI-SA-ES-NSSI or 1-0-0-0) that are actually part of the true label set (e.g., +SI+SA-ES-NSSI or 1-1-0-0), leading to false negatives. Both hallucination and omission errors (hybrid errors) can occur simultaneously, where the model predicts some incorrect labels while missing others in predicted label set (e.g., SA and NSSI in +SI-SA-ES+NSSI or 1-0-0-1), compared with the true label set (e.g., +SI+SA-ES-NSSI or 1-1-0-0).

**Acknowledgements**
Research reported in this publication was supported by the National Library of Medicine of the National Institutes of Health under award numbers R01LM011934, the National Human Genome Research Institute

**Reference**

1. Organization, W.H. *Suicide worldwide in 2019: global health estimates*, (2021).
2. World Health Organization. Suicide. (2024).
3. Organization, W.H. *Preventing suicide: A global imperative*, (2014).
4. Centers for Disease Control and Prevention. Suicide Data and Statistics. (2024).
5. Hawton, K. & Pirkis, J. Preventing suicide: a call to action. *The Lancet Public Health* **9**, e825-e830 (2024).
6. Chan, M.K.Y.*, et al*. Predicting suicide following self-harm: systematic review of risk factors and risk scales. *British Journal of Psychiatry* **209**, 277-283 (2016).
7. Hill, N.T.*, et al*. Association of suicidal behavior with exposure to suicide and suicide attempt: A systematic review and multilevel meta-analysis. *PLoS medicine* **17**, e1003074 (2020).
8. McHugh, C.M., Corderoy, A., Ryan, C.J., Hickie, I.B. & Large, M.M. Association between suicidal ideation and suicide: meta-analyses of odds ratios, sensitivity, specificity and positive predictive value. *BJPsych Open* **5**, e18 (2019).
9. Cusick, M.*, et al*. Using weak supervision and deep learning to classify clinical notes for identification of current suicidal ideation. *Journal of psychiatric research* **136**, 95-102 (2021).
10. Cusick, M.*, et al*. Portability of natural language processing methods to detect suicidality from clinical text in US and UK electronic health records. *Journal of affective disorders reports* **10**, 100430 (2022).
11. Wang, X.*, et al*. Comparisons of deep learning and machine learning while using text mining methods to identify suicide attempts of patients with mood disorders. *Journal of affective disorders* **317**, 107-113 (2022).
12. Rawat, B.P.S., Kovaly, S., Pigeon, W.R. & Yu, H. Scan: Suicide attempt and ideation events dataset. in *Proceedings of the conference. Association for Computational Linguistics. North American Chapter. Meeting*, Vol. 2022 1029 (NIH Public Access, 2022).
13. Xie, F., Ling Grant Deborah, S., Chang, J., Amundsen Britta, I. & Hechter Rulin, C. Identifying Suicidal Ideation and Attempt From Clinical Notes Within a Large Integrated Health Care System. *The Permanente Journal* **26**, 85-93 (2022).
14. Nordin, N., Zainol, Z., Mohd Noor, M.H. & Chan, L.F. Suicidal behaviour prediction models using machine learning techniques: A systematic review. *Artificial Intelligence in Medicine* **132**, 102395 (2022).
15. Singhal, K.*, et al*. Large language models encode clinical knowledge. *Nature* **620**, 172-180 (2023).

16. Thirunavukarasu, A.J.*, et al*. Large language models in medicine. *Nature Medicine* **29**, 1930-1940 (2023).
17. Li, Z.*, et al*. Suicide Phenotyping from Clinical Notes in Safety-Net Psychiatric Hospital Using Multi-Label Classification with Pre-Trained Language Models. in *AMIA Summits Transl Sci Proc*, Vol. In press (American Medical Informatics Association, Pittsburgh, PA, USA, 2025).
18. Minaee, S.*, et al*. Large language models: A survey. *arXiv preprint arXiv:2402.06196* (2024).
19. Zandbiglari, K., Kumar, S., Bilal, M., Goodin, A. & Rouhizadeh, M. Enhancing suicidal behavior detection in EHRs: A multi-label NLP framework with transformer models and semantic retrieval-based annotation. *Journal of Biomedical Informatics* **161**, 104755 (2025).
20. Keloth, V.K.*, et al*. Social determinants of health extraction from clinical notes across institutions using large language models. *npj Digital Medicine* **8**, 1-13 (2025).
21. Van Orden, K.A.*, et al*. The interpersonal theory of suicide. *Psychological review* **117**, 575 (2010).
22. Chu, C.*, et al*. The interpersonal theory of suicide: A systematic review and meta-analysis of a decade of cross-national research. *Psychological bulletin* **143**, 1313 (2017).
23. Cohan, A.*, et al*. SMHD: a Large-Scale Resource for Exploring Online Language Usage for Multiple Mental Health Conditions. 1485-1497 (Association for Computational Linguistics, Santa Fe, New Mexico, USA, 2018).
24. Zhang, M.-L. & Zhou, Z.-H. A review on multi-label learning algorithms. *IEEE transactions on knowledge and data engineering* **26**, 1819-1837 (2013).
25. Liu, W., Wang, H., Shen, X. & Tsang, I.W. The emerging trends of multi-label learning. *IEEE transactions on pattern analysis and machine intelligence* **44**, 7955-7974 (2021).
26. Bogatinovski, J., Todorovski, L., Džeroski, S. & Kocev, D. Comprehensive comparative study of multi-label classification methods. *Expert Sys Appl* **203**, 117215 (2022).
27. Sulieman, L.*, et al*. Classifying patient portal messages using Convolutional Neural Networks. *Journal of biomedical informatics* **74**, 59-70 (2017).
28. Wu, X., Zhang, J. & Li, H. Text-to-table: A new way of information extraction. *arXiv preprint arXiv:2109.02707* (2021).
29. Nasar, Z., Jaffry, S.W. & Malik, M.K. Named entity recognition and relation extraction: State-of-the-art. *ACM Computing Surveys (CSUR)* **54**, 1-39 (2021).
30. Wang, Y.*, et al*. Clinical information extraction applications: a literature review. *Journal of biomedical informatics* **77**, 34-49 (2018).
31. Fu, S.*, et al*. Clinical concept extraction: a methodology review. *Journal of biomedical informatics* **109**, 103526 (2020).
32. Hu, Y.*, et al*. Improving large language models for clinical named entity recognition via prompt engineering. *J Am Med Inform Assoc* **31**, 1812-1820 (2024).
33. Dagdelen, J.*, et al*. Structured information extraction from scientific text with large language models. *Nature Communications* **15**, 1418 (2024).

34. Fu, S. University of Minnesota (2021).
35. Sun, S. Meta-analysis of Cohen's kappa. *Health Services and Outcomes Research Methodology* **11**, 145-163 (2011).
36. Brown, T.B. Language models are few-shot learners. *arXiv preprint arXiv:2005.14165* (2020).
37. Achiam, J*., et al*. Gpt-4 technical report. *arXiv preprint arXiv:2303.08774* (2023).